\documentclass[11pt,a4paper]{article}
\usepackage[hyperref]{acl2021}
\usepackage{times}
\usepackage{latexsym}

\usepackage{booktabs}
\usepackage[T1]{fontenc}
\usepackage{amssymb}
\usepackage{pifont}
\usepackage{url}
\usepackage{multirow}
\usepackage{graphicx}
\usepackage{bm}
\usepackage{subfigure}
\usepackage{soul}
\usepackage{color}
\usepackage{comment}
\newcommand{\hlred}{\colorlet{c}{red!20}\sethlcolor{c}\hl}
\newcommand{\hlgreen}{\colorlet{c}{green!20}\sethlcolor{c}\hl}
\newcommand{\hlyellow}{\colorlet{c}{yellow!20}\sethlcolor{c}\hl}

\usepackage{microtype}

\aclfinalcopy 
\def\aclpaperid{433} 

\title{OpenMEVA: A Benchmark for Evaluating Open-ended Story Generation Metrics}

\author{
 Jian Guan$^1$, Zhexin Zhang$^1$, Zhuoer Feng$^1$, Zitao Liu$^2$, Wenbiao Ding$^2$, \\\textbf{Xiaoxi Mao$^3$, Changjie Fan$^3$ and Minlie Huang}$^1$\Thanks{~Corresponding author}\\
\small{$^1$The CoAI group, DCST; $^1$Institute for Artificial Intelligence; $^1$State Key Lab of Intelligent Technology and Systems;}\\
\small{$^1$Beijing National Research Center for Information Science and Technology;} 
\small{$^1$Tsinghua University, Beijing 100084, China.}\\
\small{$^2$TAL Education Group.  $^3$Netease Fuxi AI Lab.}\\
  \small{\texttt{\{j-guan19,zx-zhang18,fze17\}@mails.tsinghua.edu.cn}, \texttt{zitao.jerry.liu@gmail.com}}, \\\small{\texttt{dingwenbiao@100tal.com}, \texttt{ \{maoxiaoxi,fanchangjie\}@corp.netease.com}}, \small{\texttt{aihuang@tsinghua.edu.cn}} \\
}
\date{}

\begin{document}
\maketitle
\begin{abstract}
Automatic metrics are essential for developing natural language generation~(NLG) models, particularly for open-ended language generation tasks such as story generation. However, existing automatic metrics are observed to correlate poorly with human evaluation. The lack of standardized 
benchmark datasets makes it difficult to fully evaluate the capabilities of a metric and fairly compare different metrics. 
Therefore, we propose OpenMEVA, a benchmark for evaluating open-ended story generation metrics. OpenMEVA provides a comprehensive test suite to assess the capabilities of metrics, including (a) the correlation with
human judgments, (b) the generalization to different model outputs and datasets, (c) the ability to judge story coherence, and (d) the robustness to perturbations. To this end, OpenMEVA includes both manually annotated stories and auto-constructed test examples. 
We evaluate existing metrics on OpenMEVA and observe that they have poor correlation with human judgments, fail to recognize discourse-level incoherence, and lack inferential knowledge~(e.g., causal order between events), the generalization ability and robustness. Our study presents insights for developing NLG models and metrics in further research.

\end{abstract}
\section{Introduction}
Significant advances have been witnessed in many NLG tasks with pretraining models~\cite{devlin2018bert,brown2020language}. 
However, existing generation models are still far behind the human-level performance to generate {reasonable} texts, particularly for open-ended generation tasks such as story generation~\cite{fan2018hierarchical,guan2020knowledge}. One critical obstacle is the lack of powerful metrics for measuring the quality of generation. 

The standard paradigm for evaluating NLG metrics is to calculate the correlation with human judgments on manually annotated datasets~\cite{tao2018ruber,sellam2020bleurt}. Recent studies have discovered that the existing automatic metrics may correlate poorly with human judgments~\cite{liu2016not,guan2020union}. 
Unfortunately, the lack of benchmark datasets 
makes it challenging to completely assess the capabilities of a metric and fairly compare different metrics. Firstly,  
annotated datasets 
usually contain innate data bias and annotation bias. Secondly, summarizing the performance with a single aggregate statistic~(e.g., a correlation score) makes it difficult to probe which aspects a metric can successfully capture and which can not. Therefore, many alternative approaches have been proposed to evaluate NLG metrics, such as measuring the robustness to adversarial examples~\cite{zhang2019bertscore}, and the generalization to quality-biased data~\cite{sellam2020bleurt}. However, these approaches only focus on an individual capability or a single task, 
thereby failing to fully reveal the strengths and weaknesses of a NLG metric.

Therefore, we propose OpenMEVA, a benchmark for \textit{\textbf{Open}-ended story generation \textbf{M}etrics \textbf{Eva}luation}. 
We first collect a  \textit{MAN}ually annotated \textit{S}tory dataset~(\textsc{mans}).   
The stories are generated by various generation models trained on two widely used story corpora, ROCStories~\cite{mostafazadeh2016corpus} and WritingPrompts~\cite{fan2018hierarchical}.
Therefore, \textsc{mans} supports to evaluate metrics in terms of not only the correlation with human judgments, but also the generalization w.r.t {\it model drift}~(generations from different models) and {\it dataset drift}~(examples from different datasets).

In addition, OpenMEVA also includes an \textit{AUTO}-cons{t}ructed \textit{S}tory dataset~(\textsc{autos}) to test the robustness and the ability to judge story coherence, namely, the semantic relations and discourse structures in the context.  
We construct \textsc{autos} by perturbing human-written stories,   
and test the metrics in each single aspect~(e.g., the ability to recognize inconsistency) by validating the input-output behavior~\cite{ribeiro-etal-2020-beyond}. 
Through such behavioral tests, \textsc{autos} can support to reveal potential issues of metrics in multiple aspects, 
which would be not traceable in machine-generated examples in \textsc{mans}.

We conduct extensive experiments to assess the capabilities of existing automatic metrics on OpenMEVA. We find that state-of-the-art metrics still correlate poorly~(less than 0.5) with human judgments on \textsc{mans}. And it is difficult for the learnable metrics to generalize to \textit{model or dataset drift}. Through tests on \textsc{autos}, we observe that most metrics can perform well in recognizing incoherence at \textit{token level}~(e.g., unrelated entities) and \textit{sentence level}~(e.g., semantic repetition), 
but fail to recognize \textit{discourse-level} incoherence~(e.g., inconsistency) and lack understanding of inferential knowledge~(e.g., temporal order between events). Besides, we also show that existing metrics are not robust to a small number of typos and synonym substitution. These findings may inspire new directions for developing NLG models and designing  metrics in future research.

    
We also provide an open-source 
toolkit
which implements various metrics, and therefore supports the comparison and analysis of metrics. In addition, the toolkit provides data perturbation techniques for generating customized test cases beyond \textsc{autos}, which can facilitate fast development of new automatic metrics\footnote{All the tools, data, and evaluation scripts are available at \url{https://github.com/thu-coai/OpenMEVA}}.

\section{Related Work}
Various automatic metrics have been proposed for evaluating language generation. They can be roughly divided into referenced, unreferenced, and hybrid metrics, according to whether relying on human-written references when calculating the metric score. Referenced metrics usually measure the similarity between a sample and some references based on word-overlap~(e.g., BLEU~\cite{papineni2002bleu}, ROUGE~\cite{lin-2004-rouge}) or word embedding~(e.g., BERTScore~\cite{zhang2019bertscore}, MoverScore~\cite{zhao2019moverscore}). However, referenced metrics were reported to correlate poorly with human judgments in open-ended generation tasks~\cite{liu2016not} due to the one-to-many issue~\cite{zhao2017learning}. 
To address the issue, unreferenced metrics were proposed to measure the quality of a sample without any reference, such as perplexity, discriminator-based metric~\cite{kannan2017adversarial},  \textsc{Union}~\cite{guan2020union} and GRADE~\cite{huang2020grade}.
Besides, hybrid metrics combine referenced and unreferenced metrics~(e.g., RUBER and its variant~\cite{tao2018ruber, ghazarian2019better}) or learn from the human-annotated score~(e.g., ADEM~\cite{lowe2017towards},  
BLEURT~\cite{sellam2020bleurt}). 

Recently, there have been many criticisms for existing metrics. \citet{garbacea2019judge} showed the poor generalization of discriminator-based metrics. 
\citet{sai2019re} demonstrated ADEM is not robust to simple attacks such as simple word substitution or random word shuffle. 
However, these criticisms only focus on individual metrics or capabilities. 
Notably, \citet{ribeiro-etal-2020-beyond} proposed a framework CheckList to evaluate different capabilities of general language understanding models by validating the input-output behavior. The test cases are created from scratch or by perturbing an existing dataset. Similar to Checklist, OpenMEVA also employs automatically constructing examples for behavioral tests. However, CheckList only focuses on single sentences, thereby lacking the ability to test models in understanding long texts with many discourse-level features~(e.g., temporal relationship). 
Moreover, the testing methods of CheckList are not directly applicable for NLG metrics. Specifically, CheckList measures the performance of a model by calculating the failure rate between discrete model prediction and automatic labels. Such failure rates are ineffective for measuring metrics since most metric scores are continuous.
To address the above issues, we propose perturbation techniques and testing methods more applicable for story generation metrics.

\begin{figure*}[!htp]
  \centering
\includegraphics[width=0.9\linewidth]{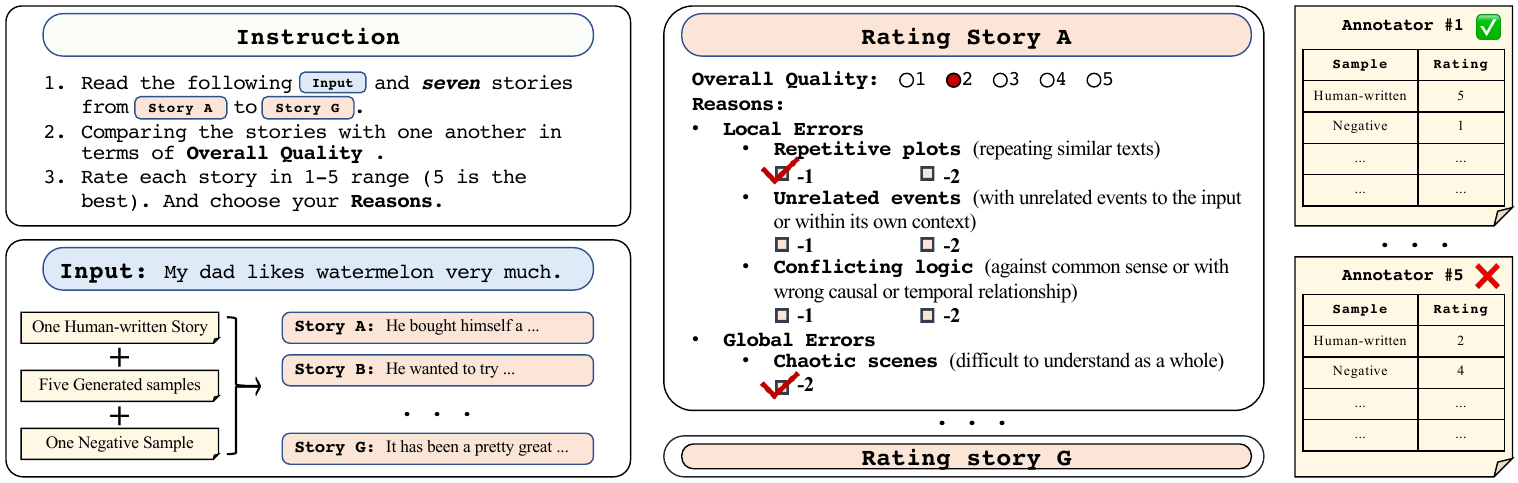}
  \caption{Overview for the manual annotation interface. 
  \texttt{Story A} gets two points in overall quality since it gets three points deducted for its repetitive plot and chaotic scene.  The ratings of \texttt{Annotator \#5} for the current story group are rejected because of the low score for the human-written story and the high score for the negative sample.}
  \label{fig:ant}
\end{figure*}

\section{Data Collection}
We collect \textsc{mans} and \textsc{autos} based on ROCStories~(\textbf{ROC} for short)~\cite{mostafazadeh2016corpus} and WritingPrompts~(\textbf{WP} for short)~\cite{fan2018hierarchical}, which are commonly used for story generation~\cite{guan2020knowledge,fan-etal-2019-strategies} and evaluation~\cite{guan2020union}. ROC contains 98,162 
five-sentence commonsense stories with about 50 words, while WP consists of 303,358 pairs of prompts and stories, which are usually unconstrained on writing topics. We retain about 250 words~(with correct sentence boundary) 
for stories in WP.
Although we only consider 
the stories in the two corpora, OpenMEVA is designed to measure the capability of NLG metrics to evaluate general linguistic features such as coherence, 
which may pertain to other stories. Besides, our idea that building datasets by manual annotation or automatic construction can be easily extended to evaluate specific aspects for other types of stories. 


\subsection{\textsc{mans}: Manually Annotated Stories}\label{sec:mags}

We collect \textsc{mans} to assess the correlation of metrics with human judgments and the generalization ability when evaluating machine-generated stories. 
We randomly split ROC and WP by 90\%/5\%/5\% for training/validation/test of the generation models. We regard the first sentence for ROC and the prompt for WP as input. After training, we generate stories based on the test sets. 
Then, we resort to Amazon Mechanical Turk~(AMT) for human judgments of the generated stories. 
We consider various generation models including a \textbf{Seq2Seq} model~\cite{sutskever2014sequence}, \textbf{Fusion} ~\cite{fan2018hierarchical}, \textbf{Plan\&Write}~\cite{yao2018plan},  the fine-tuned \textbf{GPT-2}~\cite{radford2019language} and \textbf{K}nowled\textbf{G}e\textbf{-}enhanced \textbf{GPT-2}~\cite{guan2020knowledge}. These models cover diverse network architectures and different levels of the generation ability, which support to evaluate the generalization to examples with different model biases or quality levels.

\paragraph{Manual Annotation}
We present the manual annotation interface in Figure~\ref{fig:ant}. In each human intelligence task (HIT) of AMT, we show workers the input of a story paired with \textit{seven} stories including (a) five stories generated by the above five models, (b) the human-written story, and (c) a negative example constructed by perturbing a story~(e.g., repetition, shuffling) sampled from the test sets. Then we ask workers to compare the \textit{overall quality} of the seven stories\footnote{We do not ask annotation in other aspects~(e.g., \textit{interesting}) since previous work~\cite{novikova2017we} has noted that the annotation scores on different aspects are highly correlated in spite of careful design. And computing correlation scores in the entangled aspects would be unconvincing. 
}, and rate each story with a 5-point Likert scale.  
We reject an HIT if the worker rates the human-written story lower than four points or rates the negative example higher than two points. Through the quality control mechanism, we filtered about 38.7\% assignments for ROC and 75.4\% for WP. 
Finally, we ensure that there are five valid 
ratings for each generated story, and we regard the average rating as the final human judgment.

\begin{table*}[!ht]
\tiny
    \centering
    \begin{tabular}{p{32pt}|p{105pt}|p{281pt}}
    \toprule
\textbf{Aspects}&\textbf{Selecting Coherent Examples}&\textbf{Creating Incoherent Examples}\\ 
\midrule
\textbf{Lexical}\newline \textbf{Repetition} & All the human-written stories.
& \textbf{(1)} Repeating a 4-gram~(with “and” inserted before it).  
\textbf{(2)} Repeating a sentence.\newline
\texttt{Case:} ... {he stepped on the stage \hlgreen{and stepped on the stage}} ...\\
\midrule
\textbf{Semantic}\newline \textbf{Repetition} & All the human-written stories.
&
\textbf{(1)} Repeating a sentence with its paraphrase by back translation\footnotemark. To ensure the semantic similarity and avoid much word overlap, we only use those paraphrases whose MoverScore is larger than 0.4 and BLEU-1 is less than 0.6 with the original sentences. We present some examples for paraphrase generation in the appendix. 
\newline
\texttt{Case:}~he hired an attorney. \hlgreen{he employed a lawyer} ... \textit{(MoverScore=0.57, BLEU-1=0.40)}
\\
\midrule
\textbf{Character}\newline \textbf{Behavior}  & Stories with passive voice or with personal pronouns~(e.g., ``him'', ``their'') 
for multiple characters.\newline\texttt{Case:} 
... \textit{it} asked {\textit{John}} if \textit{John} could ...
&
\textbf{(1)} Reordering the subject and object of a sentence. \textbf{(2)} Substituting a personal pronoun with another one which refers to other characters. And we do no change the grammatical case of the substituted pronoun 
(e.g., ``my'' can be substituted with ``his'' but not with ``him'').
\newline\texttt{Case:} 
\hlyellow{\textit{John}} $\leftrightarrow$ asked \hlyellow{\textit{it}} if \textit{John} could ...\\ 
\midrule
\textbf{Common}\newline \textbf{Sense}& 
Stories with both the head and tail entities of a triple in ConceptNet\protect\footnotemark\cite{speer2012representing}.
&
\textbf{(1)} Substituting 10\% entities with its neighboring entity in ConceptNet.\newline\texttt{Case:}~today is \hlred{\textit{Halloween}} $\mapsto$ \hlgreen{\textit{Christmas}} . Jack is excited to go \textit{trick or treating} ... (``Halloween'' and ``{Christmas}'' has the relation ``{Antonyms}'')\\
\midrule
\textbf{Consistency} & Stories with negated words~(e.g., ``not'', ``hardly'', ``inactive'').\newline\texttt{Case:} 
... Tom decided \textit{not} to give up ...
&
\textbf{(1)} Substituting words with the antonyms~(e.g., ``happy'' vs. ``upset''), which are retrieved from WordNet~\cite{miller1998wordnet}. The antonyms are converted to the same form 
(e.g., verb tense) with the original words. 
\textbf{(2)} Inserting or Deleting negated words for 20\% sentences. 
\newline\texttt{Case:} she \hlred{\textit{agreed}} $\mapsto$ \hlgreen{\textit{disagreed}} to get vaccinated ... \\
\midrule
\textbf{Relatedness}  & Stories with weak token-level semantic relatedness within the context\footnotemark.
\newline\texttt{Case:} Craig was diagnosed with cancer. he decided to fight it ... 
&\textbf{(1)} Substituting 25\% nouns or verbs randomly (with correct word forms). \textbf{(2)} Substituting a sentence randomly with another sampled from the dataset.\newline\texttt{Case:}
Craig was diagnosed with cancer. \hlred{he decided to fight it.} $\mapsto$ \hlgreen{Kelly wanted to put up the Christmas tree.} He tried several different approaches and medications. eventually it went into remission ... \\
\midrule
\textbf{Causal}\newline \textbf{Relationship} &Stories with causality-related words~(e.g., ``because'').\newline\texttt{Case:} ... {the sky is clear.} \textit{so} he can see it .
&\textbf{(1)} Reordering the cause and effect, which should be two individual sentences or two clauses connected by a causality-related conjunction
; \textbf{(2)} Substituting the causality-related words with the antonyms (e.g., ``reason'' vs. ``result''). \newline\texttt{Case:} ... \hlyellow{he can see it.} $\leftrightarrow$ \textit{so} \hlyellow{the sky is clear}.\\
\midrule
\textbf{Temporal}\newline \textbf{Relationship} & Stories with time-related words (e.g., ``before'',``then'').\newline\texttt{Case:} 
... Tina \textit{then} learnt her lesson.&
\textbf{(1)} Reordering two sequential events, which should be two individual sentences or two clauses connected by a time-related conjunction. \textbf{(2)} Substituting the time-related words with the antonyms (e.g., ``after'' vs. ``before''). \newline\texttt{Case:}  ... \hlred{after} $\mapsto$ \hlgreen{before} eating one bite I was not hungry. \\
\bottomrule
    \end{tabular}
    \caption{Examples for the discrimination test to evaluate the ability to judge story coherence in different \textbf{aspects}. \textit{Italic} words indicate the crucial keywords for the corresponding aspects. The \textbf{coherent examples} are selected from the human-written stories. The \textbf{incoherent examples} are created by perturbation including \hlgreen{insertion}, \hlred{deletion} and \hlyellow{reordering}, where \textbf{(1)} and \textbf{(2)} mean different perturbation techniques.}
    \label{tab:judgment}
\end{table*}

\begin{table}[!ht]
\tiny
    \centering
    \begin{tabular}{p{32pt}|p{163pt}}
    \toprule
    \textbf{Aspects} & \textbf{Perturbations}\\
    \midrule
    \textbf{Synonyms} & Substituting a word with its synonym retrieved from WordNet.
    \newline\texttt{Case:} ... I \hlred{purchased} $\mapsto$ \hlgreen{bought} my uniforms.\\
    \midrule
    \textbf{Paraphrases} &  Substituting a sentence with its paraphrase. 
    \newline\texttt{Case:} 
    \hlred{he hired an attorney} $\mapsto$ \hlgreen{he employed a lawyer} \\
    \midrule
    \textbf{Punctuation} & Deleting inessential punctuation marks~(e.g., commas).
    \newline\texttt{Case:} ... eventually\hlred{,} he became hungry ...\\
    \midrule
    \textbf{Contraction} & Contracting or Expanding contraction.\newline\texttt{Case:} ... I\hlred{'ll} $\mapsto$ \hlgreen{ will} have to keep waiting ...\\
    \midrule
    \textbf{Typos} &  Swapping two adjacent characters;  Repeating or Deleting a character. We modify less than 2\% words of an example to avoid much noise. \newline\texttt{Case:} ... an \hlred{orange} $\mapsto$ \hlgreen{ornage} broke her nose.\\
    \bottomrule
    \end{tabular}
    \caption{Examples for the invariance test to evaluate the {robustness} to \textbf{perturbations} in different \textbf{aspects}.}
    \label{tab:robust_case}
\end{table}

Considering that overall quality is often too abstract to measure, we follow 
previous recommendations~\cite{belz2014towards,van2020human} to decide the overall quality by summarizing multiple separate criteria. 
We ask the workers to decide the rating of a story based on a point deduction policy. 
Specifically, a story should get punishment in points if it contains errors such as \textit{repetitive plots}, \textit{unrelated events} and \textit{conflicting logic}, or globally \textit{chaotic scenes}, which are commonly observed in existing NLG models~\cite{guan2020union}~(several examples shown in the appendix). Intuitively, the policy can alleviate the tendency to give high scores and ensure that the judgment standard of workers is as consistent as possible during annotation. 
To avoid introducing extra bias in the policy, we do not impose the restriction on workers to exactly match the rating in overall quality with the deducted points.

\paragraph{Data Statistics}
We randomly sampled 200 stories from test sets of ROC and WP for story generation, respectively. Therefore, \textsc{mans} contains $2\times200\times5=2,000$ annotated machine-generated stories, paired with corresponding inputs and human-written references. The Krippendorff's $\alpha$~\cite{krippendorff2018content} of the human judgments is $0.77/0.71$ for ROC/WP, indicating a moderate inter-annotator agreement ($\alpha\in[0.67,0.8]$). 
We show more statistical details in the appendix. 

\subsection{\textsc{autos}:  Auto-Constructed Stories}

While improving correlation with human judgments is the ultimate goal for developing automatic metrics, merely relying on limited annotated data may make the true evaluation performance overestimated~\cite{ribeiro-etal-2020-beyond}. Besides, a machine-generated story may contain multiple entangled errors~(e.g., repetition, unrelatedness),  
which do not support individual tests for metrics. Therefore, we propose to evaluate the capabilities of metrics with auto-constructed test examples~(i.e., \textsc{autos}), each of which is created to focus on a single aspect. We construct \textsc{autos} based on the human-written stories in  the test sets of ROC and WP. 

\footnotetext[2]{We generate paraphrases based on the 
back translation augmentation system of UDA~\cite{xie2020unsupervised}.}
\footnotetext[3]{ConceptNet is a knowledge base including millions of commonsense triples like (\texttt{h}, \texttt{r}, \texttt{t}), meaning that the head entity \texttt{h} has a relation \texttt{r} with the tail entity \texttt{t}. Note that we only regard nouns and verbs as entities.}
\footnotetext[4]{We regard the stories with maximum inter-sentence MoverScore less than 0.1 as those which have weak token-level semantic relatedness within the context.}

\paragraph{Aspects}
We argue that an ideal metric for evaluating open-ended language generation should have {at least} the following capabilities: (a)~ the ability to \textit{judge} story coherence, which requires recognizing \textbf{lexical} and \textbf{semantic repetition}, 
unreasonable \textbf{character behavior}~(e.g., chaotic coreferences), 
violation of \textbf{common sense}~(e.g., \textit{``trick or treat''} on \textit{``Christmas''}), 
poor \textbf{consistency} 
and \textbf{relatedness}, 
incorrect \textbf{causal} and \textbf{temporal relationship}; and (b)~the \textit{robustness} to perturbations,  
such as substituting with \textbf{synonyms} or \textbf{paraphrases}, deleting
unimportant \textbf{punctuation} marks, contracting full expressions or expanding \textbf{contractions}, and adding \textbf{typos}. 
Tests in these aspects require metrics to fully understand the linguistic features at token level~(e.g., synonyms), sentence level~(e.g., semantic similarity), and discourse level~(e.g., context relatedness in content and proper sentence orders), 
and possess knowledge about common sense, causality, etc., which are usually not  traceable in machine-generated stories. 
Although these aspects are not exhaustive, it is a starting point for further research.  Table~\ref{tab:judgment} and \ref{tab:robust_case} present some examples for the two capabilities, respectively. 

\paragraph{Test Types}
We create examples with different test types to evaluate the above capabilities of metrics. Firstly, we evaluate the ability to judge story coherence by the \textit{discrimination test}, which requires metrics to distinguish human-written coherent examples from incoherent ones. We create each incoherent example by applying perturbation within a single aspect. 
Besides, we also select different human-written stories as coherent examples for different aspects, as shown in Table~\ref{tab:judgment}. 
For robustness assessment, we expect the metric scores to remain the same with certain perturbations, i.e., the \textit{invariance test}, as shown in Table~\ref{tab:robust_case}.

However, the perturbation may inevitably introduce grammar errors. 
To alleviate the issue, 
we filter out those ungrammatical examples in \textsc{autos} except for those used to evaluate robustness to typos using an automatic grammaticality classifier. 
We present the statistics of \textsc{autos} together with the evaluation results in Table~\ref{tab:acts}/ \ref{tab:robust} for the discrimination/invariance tests, respectively. And we provide more details about the construction of \textsc{autos} and the grammaticality classifier in the appendix.

\section{Evaluation}

\begin{table*}[!ht]
\scriptsize
    \centering
    \begin{tabular}{l|c|cccc|c|cccc}
    \toprule
\multirow{4}{*}{\textbf{Metrics}} & \multicolumn{5}{c|}{\textbf{ROC}}&\multicolumn{5}{c}{\textbf{WP}}\\
&\multirow{2}{*}{\textbf{Overall}}&\multicolumn{4}{c|}{\textbf{46 Reasonable Samples +}}&\multirow{2}{*}{\textbf{Overall}}&\multicolumn{4}{c}{\textbf{35 Reasonable Samples +}}\\
&&\multicolumn{1}{c}{\textbf{Rept}} &\multicolumn{1}{c}{\textbf{Unrel}} &\multicolumn{1}{c}{\textbf{Conf}}&\multicolumn{1}{c|}{\textbf{Chao}}&&\multicolumn{1}{c}{\textbf{Rept}} &\multicolumn{1}{c}{\textbf{Unrel}} &\multicolumn{1}{c}{\textbf{Conf}}&\multicolumn{1}{c}{\textbf{Chao}}\\ 
&\textbf{1,000}&\textbf{22}&\textbf{319}&\textbf{39}&\textbf{87}&\textbf{1,000}&\textbf{23}&\textbf{330}&\textbf{83}&\textbf{24}\\
	\midrule
	\midrule
    \texttt{BLEU}&-0.0239~~~&0.0520~~~&0.0192~~~&0.1134&0.0156~~~&-0.0537~~~&0.1188&-0.0421~~~~~&-0.0875~~~&-0.1451~~~~~\\
    \texttt{BERTScore-F1}&~0.1271$^{*}$&0.1396~~~&0.1240~~~&0.0626~~~&0.2283$^{*}$&~0.0329~~~&0.1198&0.0446~~~&0.0189&0.0634~~~\\
		\midrule
	\texttt{PPL (P)}&~0.2547$^{*}$&-0.1075~~~~&0.1105~~~&0.1354&0.5248$^{*}$&~0.3033$^{*}$&0.0219&\textbf{0.1853}$^{*}$&\textbf{0.2188}&\textbf{0.4428}$^{*}$\\
	\texttt{PPL (F)}&~0.2817$^{*}$&0.2152~~~&0.1380$^{*}$&\textbf{0.2643}&\textbf{0.5910}$^{*}$&~0.2952$^{*}$&0.0179&0.1720$^{*}$&0.1917&0.3182$^{*}$\\
	\texttt{R$_u$-BERT}&~0.0830$^{*}$&0.1160~~~&0.0877~~~&0.1103&0.1774~~~&0.1666$^{*}$&0.0936&0.0793~~~&0.0162&0.0077~~~\\	
    {\texttt{\textsc{Union}}}&~\textbf{0.4119}$^{*}$&\textbf{0.4517}$^{*}$&\textbf{0.2000}$^{*}$&0.2107&0.4695$^{*}$&~\textbf{0.3256}$^{*}$&\textbf{0.3283}&0.1738$^{*}$&0.1914&0.3967$^{*}$\\
	\midrule
    \texttt{RUBER-BERT}&~{0.1434}$^{*}$&0.0813~~~&0.1453$^{*}$&0.1173&0.1723~~~&0.2116$^{*}$&0.0716&0.1132~~~&0.0721&0.1493~~~  \\
	\bottomrule
    \end{tabular}
    \caption{Pearson correlation with human judgments on \textsc{mans}. 
    \texttt{PPL~(P)} and \texttt{PPL~(F)} mean Perplexity estimated by \textit{pretrained} and  \textit{fine-tuned} GPT-2, respectively. The best performance 
    is highlighted in \textbf{bold}. 
    The results contain the correlation with human judgments on all the annotated samples in \textsc{mans}~(\textbf{Overall}), and the correlation with the binary labels on reasonable samples and unreasonable ones of different error types. The error types include \textbf{Repe}titive plots, \textbf{Unrel}ated events, \textbf{Conf}licting logic and \textbf{Chao}tic scenes. The numbers in the table header denote the number of corresponding stories. * indicates the correlation score is significant (p-value<0.01).}
    \label{tab:corr}
\end{table*}

\begin{figure*}[!htp]
  \centering
\includegraphics[width=\linewidth]{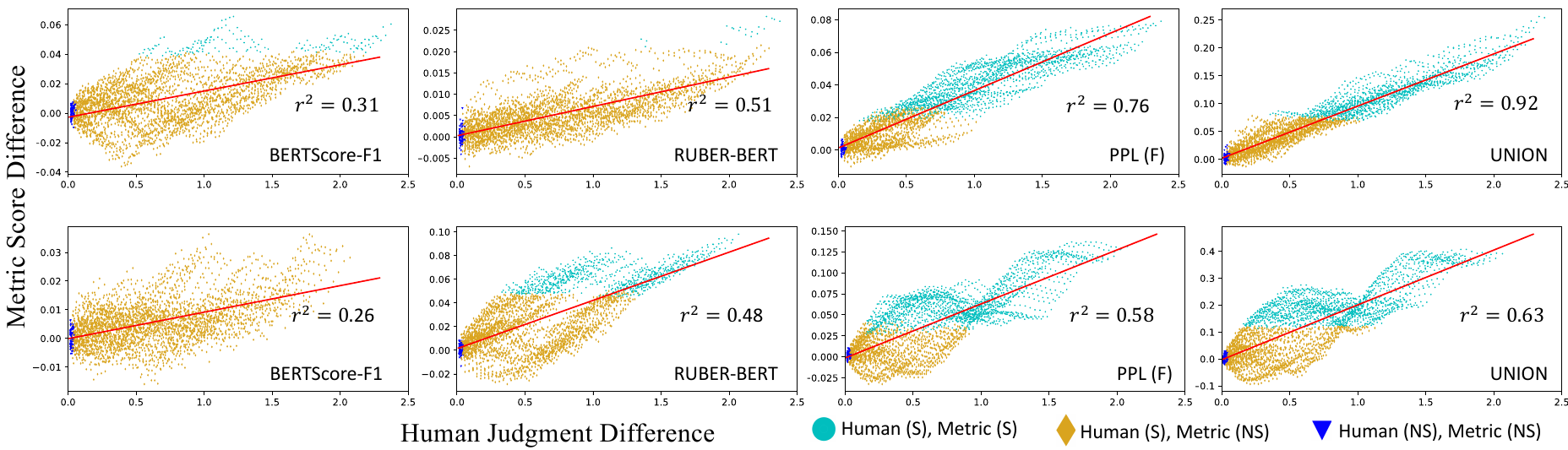}
  \caption{Correlation between human judgment difference~(x-axis) and metric score difference~(y-axis). Top: ROC, Bottom: WP. We only show the situation in the positive x-axis, since it is centrosymmetric with that in the negative x-axis. \textbf{Human~(S)}/\textbf{Metric~(S)} means the difference of human judgment/metric score is significant~(p$<$0.01, t-test), while~\textbf{(NS)} means insignificant difference. $r^2$ is the coefficient of determination for linear regression~(red line), and is exactly the square of the Pearson correlation coefficient between the x-axis and y-axis.}
  \label{fig:reg}
\end{figure*}

We evaluated existing metrics on OpenMEVA, 
and 
analyzed the strengths and weaknesses with extensive experiments.

\subsection{Evaluated Metrics} 
We experimented with existing metrics of different types as follows: \textbf{(a) Referenced Metrics:} the word-overlap based metric sentence \texttt{BLEU} score (geometric mean from 1-gram to 4-gram)~\cite{papineni2002bleu}, the contextualized embedding based metrics, \texttt{BERTScore-F1}~\cite{zhang2019bertscore}.
\textbf{(b) Unreferenced Metrics:} \texttt{Perplexity}\footnote{We follow \citet{guan2020union} to take the minus of perplexity to ensure a higher value means better quality.} estimated by GPT-2~\cite{radford2019language} (including \texttt{pretrained} GPT-2 and GPT-2 \texttt{fine-tuned} on the training sets); the self-supervised metric \texttt{\textsc{Union}}~\cite{guan2020union}.
\textbf{(c) Hybrid Metrics:} 
\texttt{RUBER-BERT}~\cite{ghazarian2019better} that improves RUBER with contextualized embeddings from BERT~\cite{devlin2018bert}. 

In addition, we also reported the performance of the unreferenced version in RUBER-BERT, denoted as \texttt{R$_u$-BERT}. And we present results with more metrics in the appendix.

\subsection{Correlation with Human Judgments}
We first calculate the Pearson 
correlation coefficient between metric scores and human judgments on \textsc{mans}. Besides, we also evaluate metrics 
on the other four evaluation sets constructed for individual error types (described in Section~\ref{sec:mags}) 
based on \textsc{mans}. Each of them contains all the reasonable samples and the unreasonable samples of some error type. 
A reasonable sample means its overall quality score larger than four points. For an unreasonable sample, we decide it is of some error type if there is only one error type annotated by at least three of five annotators. We assign the reasonable and unreasonable samples with binary labels 1 and 0, respectively, and calculate the correlation between metric scores and the binary labels on the four evaluation sets. 


We summarize the correlation results in Table~\ref{tab:corr}. As previous studies~\cite{guan2020union} observed, 
unreferenced metrics are more competitive for evaluating open-ended language generation than referenced ones. 
PPL~(F) performs better than PPL~(P) 
on ROC but not on WP, which may be because stories in ROC are created artificially and hence differ from the general language distribution during pretraining GPT-2. Furthermore, measuring input-output relatedness~(R$_u$-BERT) is not enough for language generation evaluation. \textsc{Union} outperforms other metrics in overall quality assessment since it learns to distinguish human-written stories from negative samples with more error types.  
Interestingly, it seems easier for the metrics to recognize surface errors~(e.g., repetitive plots) or serious global errors~(e.g., chaotic scenes). \textbf{However, the best correlation with human judgments is still fairly low, and it is difficult to recognize unrelatedness and conflicting plot}. The results indicate the huge room to improve the metrics.

To further examine to what extent the improvement in an automatic metric corresponds to the improvement in human judgments, we calculate the correlation between human judgment difference and metric score difference~\cite{mathur-etal-2020-tangled}. Specifically, we sort the 1,000 stories~(for ROC and WP, respectively) in \textsc{mans} by the human judgments, and then select consecutive 200 stories from the beginning and repeat the selection with a stride 10. We finally get $(1,000-200)/10=80$ story sets\footnote{We do not construct the sets by randomly sampling since it would be difficult to cover wide enough quality levels.}.  
We decide the human judgment or metric score of each set by averaging that of the stories in the set.
We calculate the human judgment difference and metric score difference between any two sets of them~($80\times80=6,400$ pairs totally), and present the correlation between the differences in Figure~\ref{fig:reg} for several typical metrics. We can see that a significant improvement in the metrics usually corresponds to a significant improvement in human judgments~(cyan/dark gray part in Figure~\ref{fig:reg}). However, both an insignificant drop and improvement in a metric could correspond to a significant improvement in human judgments. And worse, the improvement in human judgments may have a wide range, which is particularly evident for BERTScore-F1 and RUBER-BERT~(yellow/light gray part in Figure~\ref{fig:reg}). \textbf{That is, if an NLG model achieves insignificantly better scores in the two metrics, it is quite possible that the model performs significantly worse in human judgments.}
The situation is improved when using PPL~(F) and \textsc{Union}, suggesting that they may be better to measure language generation.


\subsection{Generalization Ability}
It is extremely important for learnable metrics to deal with \textit{model drift} and \textit{dataset drift} ~\cite{garbacea2019judge,sellam2020bleurt}. Specifically, a generalizable metric should be able to evaluate different NLG models since the generation quality or inductive bias can vary significantly across models. Besides, 
we also expect a metric to reliably evaluate output from different datasets even without re-training. Therefore, we assess the generalization ability of learnable metrics, including PPL~(F), {R$_u$-BERT} and {\textsc{Union}}, which are fine-tuned on the training sets of ROC and WP, respectively.

To assess the generalization to model drift, we test the metrics on stories generated by five aforementioned models in \textsc{mans}, respectively~(200 stories by each model). 
Table~\ref{tab:gen} presents the performance, which varies considerably with models. 
R$_u$-BERT only achieves a good correlation on those stories with poor relatedness (e.g., Seq2Seq on WP). PPL~(F) and \textsc{Union} perform comparably but neither do well in evaluating all the NLG models. 

\begin{table}[!ht]
\scriptsize
    \centering
    \begin{tabular}{ll|lllll}
    \toprule
&\multirow{1}{*}{\textbf{Metrics}}&\textbf{S2S}&\textbf{P\&W}&\textbf{Fusion}&\textbf{GPT-2}&\textbf{KG-G}\\
\midrule
\multirow{3}{*}{\rotatebox{90}{ROC}}&\texttt{PPL~(F)}&~\textbf{0.14}&~0.22&~\textbf{0.12}&~0.14&~0.25$^{*}$\\
&\texttt{R$_u$-BERT}&-0.02&-0.08&~0.04&~0.12&~0.06\\
&\texttt{\textsc{Union}}&~0.12&~\textbf{0.28}$^{*}$&~0.10&~\textbf{0.15}$^{*}$&~\textbf{0.32}$^{*}$\\
\midrule
\multirow{3}{*}{\rotatebox{90}{WP}}&\texttt{PPL~(F)}&~0.11&~\textbf{0.15}&~0.05&~\textbf{0.12}&~0.13\\
&\texttt{R$_u$-BERT}&~\textbf{0.18}$^{*}$&~0.08&~0.14&~0.07&~{0.02}\\
&\texttt{\textsc{Union}}&~0.09&~0.02&~\textbf{0.15}&~0.04&~\textbf{0.15}$^{*}$\\
\bottomrule
    \end{tabular}
    \caption{Pearson correlation with human judgments to assess generalization to output from different models including Seq2Seq (S2S), Plan\&Write (P\&W), Fusion, GPT-2, KG-GPT-2~(KG-G). The best performance among the metrics is highlighted in \textbf{bold}.}
    \label{tab:gen}
\end{table}

To assess the generalization to dataset drift, we first trained the metrics on ROC and then directly used them to evaluate stories from WP, and vice versa. As shown in Table~\ref{tab:gen_data}, all the metrics drops significantly in correlation when used for the other dataset due to the difference in length and topic. PPL~(F) and \textsc{Union} also have similar performance drops but are more generalizable. \textbf{The results suggest existing metrics fall short of generalization.} 

\begin{table}[!ht]
\scriptsize
    \centering
    \begin{tabular}{l|cc|cc}
    \toprule

\multirow{2}{*}{\textbf{Metrics}}&\multicolumn{2}{c|}{\textbf{Train: ROC}}&\multicolumn{2}{c}{\textbf{Train: WP}}\\
&\textbf{Test: ROC}&\textbf{Test: WP}&\textbf{Test: ROC}&\textbf{Test: WP}\\
\midrule
\midrule
\texttt{PPL(F)}&\textbf{0.2817}$^{*}$&0.2423$^{*}$&0.2470$^{*}$&\textbf{0.2952}$^{*}$\\
\texttt{R$_u$-BERT}&\textbf{0.0830}$^{*}$&0.0379~~~&0.0891$^{*}$&\textbf{0.1666}$^{*}$\\
\texttt{\textsc{Union}}&\textbf{0.4119}$^{*}$&0.2287$^{*}$&0.2128$^{*}$&\textbf{0.3256}$^{*}$\\

    \bottomrule
    \end{tabular}
    \caption{Pearson correlation with human judgments to assess {generalization} to samples from different datasets.  
    The best performance between two test datasets~(each row) for each metric is highlighted in \textbf{bold}.}
    \label{tab:gen_data}
\end{table}

\begin{table*}[!ht]
\scriptsize
    \centering
    \begin{tabular}{rlcccccccc}
    \toprule
\multicolumn{2}{l}{ \multirow{2}{*}{\textbf{Metrics}}}&
\multicolumn{1}{c}{\textbf{Lexical}}&\multicolumn{1}{c}{\textbf{Semantic}}&\multicolumn{1}{c}{\textbf{Character}}&\multicolumn{1}{c}{\textbf{Common}}&{\multirow{2}{*}{\textbf{Consistency}}}&{\multirow{2}{*}{\textbf{Relatedness}}}&\multicolumn{1}{c}{\textbf{Causal}}&\multicolumn{1}{c}{\textbf{Temporal}}\\
&&\multicolumn{1}{c}{\textbf{Repetition}}&\multicolumn{1}{c}{\textbf{Repetition}}&\multicolumn{1}{c}{\textbf{Behavior}}&\multicolumn{1}{c}{\textbf{Sense}}&&&\multicolumn{1}{c}{\textbf{Relationship}}&\multicolumn{1}{c}{\textbf{Relationship}}\\
\midrule
\midrule
\multirow{2}{*}{\textbf{ROC}}& {\textbf{Cohe}}&4,736&4,736&1,022&1,921&455&563&476&2,376\\
&{\textbf{Incohe}}&4,049&3,243&266&448&3,666&3,570&410&1,799\\
\midrule
\multicolumn{2}{l}{ {\texttt{PPL~(P)}}}&-0.1886$^{*}$~~&-0.0719$^{*}$~~&0.2547$^{*}$&\textbf{0.4246}$^{*}$ &0.1357$^{*}$&0.0744$^{*}$&0.1002$^{*}$&0.1759$^{*}$\\
\multicolumn{2}{l}{{\texttt{PPL~(F)}}}&0.0287$^{*}$&{0.2315}$^{*}$&\textbf{0.3595}$^{*}$&{0.3976}$^{*}$&0.1630$^{*}$&0.1458~~~&\textbf{0.1568}$^{*}$&\textbf{0.2007}~~~\\
\multicolumn{2}{l}{{\texttt{R$_u$-BERT}}}&0.0121~~~&0.0543$^{*}$&0.0671$^{*}$&0.0478$^{*}$&0.0194$^{*}$&0.0764$^{*}$&-0.0075~~~~&0.0135$^{*}$\\
\multicolumn{2}{l}{{\texttt{\textsc{Union}}}}&\textbf{0.5454}$^{*}$&\textbf{0.5631}$^{*}$&{0.3191}$^{*}$&~{0.3965}$^{*}$&\textbf{0.1676}$^{*}$&\textbf{0.2045}$^{*}$&~{0.1425}$^{*}$&{0.1769}$^{*}$\\
\midrule
\midrule
\multirow{2}{*}{\textbf{WP}}&{\textbf{Cohe}}&9,922&9,922&3,911&2,052&2,914&497&4,552&9,408\\
&{\textbf{Incohe}}&9,022&8,381&173&235&6,239&851&3,057&7,092\\
\midrule
\multicolumn{2}{l}{{\texttt{PPL~(P)}}}&-0.0886$^{*}$~~&-0.0461$^{*}$~~&0.2077$^{*}$&{0.4782}$^{*}$&0.2575$^{*}$&0.1328$^{*}$&0.0355$^{*}$&0.0763$^{*}$\\
\multicolumn{2}{l}{{\texttt{PPL~(F)}}}&-0.0467$^{*}$~~&{0.0986}$^{*}$&0.2783$^{*}$&\textbf{0.4871}$^{*}$&{0.3420}$^{*}$&\textbf{0.2297}$^{*}$&\textbf{0.1597}$^{*}$&\textbf{0.1788}$^{*}$\\
\multicolumn{2}{l}{{\texttt{R$_u$-BERT}}}&0.0098~~~&0.0108~~~&-0.0299~~~~&-0.0183~~~~~&0.0137~~~&0.0054~~~&-0.0143~~~~~&0.0042~~~\\
\multicolumn{2}{l}{{\texttt{\textsc{Union}}}}&\textbf{0.2302}$^{*}$&\textbf{0.2150}$^{*}$&\textbf{0.3044}$^{*}$&0.3940$^{*}$&\textbf{0.3661}$^{*}$&0.2107$^{*}$&0.0514$^{*}$&0.0459$^{*}$\\

	\bottomrule
    \end{tabular}
    \caption{Pearson correlation with automatic labels on the discrimination test set of \textsc{autos}. The higher correlation indicates the better ability to judge story coherence in different aspects. The best performance is highlighted in \textbf{bold}. \textbf{Cohe} and \textbf{Incohe} stand for the number of coherent and incoherent examples, respectively.}
    \label{tab:acts}
\end{table*}

\begin{table*}[!ht]
\scriptsize
    \centering
    \begin{tabular}{l|cc|cc|cc|cc|cc}
    \toprule
\multirow{2}{*}{\textbf{Metrics}}&
\multicolumn{2}{c|}{\textbf{Synonym}}&\multicolumn{2}{c|}{\textbf{Paraphrase}}&\multicolumn{2}{c|}{\textbf{Punctuation}}&\multicolumn{2}{c|}{\textbf{Contraction}}&\multicolumn{2}{c}{\textbf{Typo}}\\
&\textbf{Human}&\textbf{Dis}&\textbf{Human}&\textbf{Dis}&\textbf{Human}&\textbf{Dis}&\textbf{Human}&\textbf{Dis}&\textbf{Human}&\textbf{Dis}\\
\midrule
\midrule
\textbf{ROC}&3,777&2,395&3,174&2,194&574&171&1,602&1,208&4,755&4,763\\
\midrule
\texttt{PPL (P)}&0.3162$^{*}$&0.2515$^{*}$&0.1450$^{*}$&{0.0916}$^{*}$&0.0922$^{*}$&\underline{0.0856}~~~&-0.0557~~&-0.0522$^{*}$~~&\underline{0.4124}$^{*}$&\underline{0.2616}$^{*}$\\
\texttt{PPL (F)}&0.3309$^{*}$&0.2521$^{*}$&0.2742$^{*}$&0.2022$^{*}$&0.1475$^{*}$&0.0996~~~&0.0504&0.0331$^{*}$&0.4540$^{*}$&0.2973$^{*}$\\
\texttt{RUBER$_u$-BERT}&\textbf{0.0307}$^{*}$&\textbf{0.0290}$^{*}$&\textbf{0.0255}~~~&\textbf{0.0263}~~~~&\textbf{0.0052}~~~&-\textbf{0.0140}~~~&\textbf{0.0064}&\textbf{0.0071}~~~&-\textbf{0.0112}~~~~~&\textbf{0.0042}~~~\\
\texttt{\textsc{Union}}&\underline{0.2187}$^{*}$&\underline{0.1169}$^{*}$&\underline{0.1112}$^{*}$&\underline{0.0399}$^{*}$&\underline{0.0818}$^{*}$&0.1375$^{*}$&\underline{0.0275}&\underline{0.0251}~~~&0.6021$^{*}$&0.4606$^{*}$\\
\midrule
\midrule
\textbf{WP}&6,961&35,90&7,881&2,576&4,535&2,287&8,731&4,522&15,073&15,082\\
\midrule
\texttt{PPL (P)}&0.2174$^{*}$&0.1822$^{*}$&0.0910$^{*}$&0.0617$^{*}$&0.2690$^{*}$&\underline{0.2178}$^{*}$&-{0.0222}$^{*}$~~~&-\underline{0.0157}~~~~~&0.3983$^{*}$&0.3885$^{*}$\\
\texttt{PPL (F)}&0.2964$^{*}$&0.1747$^{*}$&0.2273$^{*}$&0.1020$^{*}$&0.3822$^{*}$&0.2515$^{*}$&0.0851$^{*}$&0.0682$^{*}$&0.4603$^{*}$&0.4043$^{*}$\\
\texttt{RUBER$_u$-BERT}&-\textbf{0.0013}~~~~~&\textbf{0.0004}~~~&\textbf{0.0000}~~~&\textbf{0.0000}~~~&-\textbf{0.0256}$^{*}$~~&-\textbf{0.0308}$^{*}$~~&-\textbf{0.0012}~~~~~&-\textbf{0.0043}~~~~~&\textbf{0.0133}~~~&\textbf{0.0154}$^{*}$\\
\texttt{\textsc{Union}}&\underline{0.1077}$^{*}$&\underline{0.0843}$^{*}$&\underline{0.0389}$^{*}$&\underline{0.0292}$^{*}$&\underline{0.2182}$^{*}$&0.2224$^{*}$&\underline{0.0185}$^{*}$&0.0173$^{*}$&\underline{0.3812}$^{*}$&\underline{0.3208}$^{*}$\\
	\bottomrule
    \end{tabular}
    \caption{Pearson correlation with automatic labels on the invariance test set of \textsc{autos}. The smaller absolute value of correlation indicates the better robustness. The best performance is highlighted in \textbf{bold} and the second best is \underline{underlined}. 
    The numbers in the \textbf{ROC}/\textbf{WP} rows indicate how many human-written stories~(\textbf{Human}) and incoherent
    samples from the discrimination test set~(\textbf{Dis}) are perturbed.}
    \label{tab:robust}
\end{table*}

\subsection{Ability to Judge Story Coherence}
We assess the ability of the unreferenced metrics\footnote{It is meaningless to evaluate referenced or hybrid metrics on \textsc{autos} since the reference text of a positive example is exactly itself, which is an unfair case for unreferenced metrics.} to judge story coherence based on the discrimination test set of \textsc{autos}. We assign each test example with a binary label~(1/0 for the coherent/incoherent example).  
Then we calculate the correlation between metric scores and the binary labels on the test examples of different aspects. The higher correlation means the better ability to judge coherence.


Table~\ref{tab:acts} presents the correlation results. We summarize the results as follows: \textbf{(1) PPL is ineffective to recognize repetition errors.} The observation is accordant with the results on \textsc{mans}~(Table~\ref{tab:corr}). 
PPL (P) even has a significantly negative correlation with labels in  lexical and semantic repetition. 
\textbf{(2) PPL (F) and \textsc{Union} have better average performance than others.} R$_u$-BERT performs worst in almost all the aspects. 
\textsc{Union} has the highest average performance by a large margin on ROC but underperforms PPL~(F) on WP, indicating the shortage of \textsc{Union} when evaluating longer stories. Besides, the results show that a powerful language model may also be a powerful evaluator~(if we can alleviate its preference for repetitive texts). \textbf{(3) Existing metrics perform well in recognizing incoherence at token and sentence levels.} 
For example, they 
seem to be able to recognize unreasonable behavior for a certain character, and possess some commonsense knowledge about entity relations. However, in this work the proposed perturbation can not fully cover all possible incoherence in these aspects, 
which would be regarded as the future work. 
\textbf{(4) The metrics still struggle to recognize discourse-level incoherence.} Specifically, it is difficult to recognize inconsistent events when we insert or delete negated words, and understand the semantic relatedness across sentences. Besides, they also lack inferential knowledge about the causal and temporal relationship. The observations are also accordant with the results in Table~\ref{tab:corr} where unrelated events and conflicting logic can not be well recognized. In conclusion,
we reveal various issues of the existing metrics by the isolating behavioral testing, while they achieve moderate correlation with human judgments on \textsc{mans}. 

\subsection{Robustness Evaluation}
A reliable metric should produce similar judgments for an example with simple perturbations or attacks in the input. Therefore, it is essential to evaluate the robustness of metrics. We test the robustness on the invariance test set of \textsc{autos}. 
We assign each example with a binary label~(1/0 for the original/perturbed example). Then, we calculate the correlation between metric scores and the binary labels. 
The original examples can be sampled either from human-written stories or from the incoherent examples in the discrimination test set.

Table~\ref{tab:robust} shows the robustness results. It is not surprising that R$_u$-BERT has the ``best robustness'' since the perturbations hardly influence the input-output relatedness. The result validates the relatedness is merely one side for evaluating NLG, but not means that it is a promising direction for developing robust metrics\footnote{We can imagine that a constant metric has the perfect robustness to any perturbations, but is useless for evaluation.}.
PPL is not robust to synonym substitution because the low-frequency words introduced by the perturbations (e.g., from \textit{``happy''} to \textit{``joyful''})  can cause significant change in PPL. \textsc{Union} has better robustness on average thanks to the robust contextualized representation of BERT. 
Furthermore, both PPL and \textsc{Union} perform better in contraction than in other aspects. 
However, they are very sensitive to a small number of typos~(less than 2\% words) because typos may bring some out-of-vocabulary words. Although the issue is common for almost all the (sub)word-based metrics, it is still important to handle typos since they are also common in human writing.

\section{Conclusion}

We present OpenMEVA, a benchmark to comprehensively assess capabilities of metrics for evaluating open-ended story generation. OpenMEVA includes test examples which are created by either annotating machine-generated stories or perturbing human-written stories in terms of each single aspect. We evaluate a number of existing metrics on OpenMEVA and analyze their performance on each capability extensively. Experiments demonstrate that existing metrics still correlate weakly with human judgments, fail to recognize discourse-level incoherence, and lack inferential knowledge, generalization and robustness. 
Our study reveals the weaknesses of existing metrics and may inspire new research on designing NLG metrics.

The datasets, data augmentation tools, and implemented metrics in this paper can facilitate further research on language generation and evaluation.

\section{Ethics Statement}
We build OpenMEVA based on two existing public story datasets ROCStories~(ROC) and WritingPrompts~(WP), which are widely used for story generation and evaluation. We resorted to Amazon Mechanical Turk~(AMT) for manual annotation of stories in \textsc{mans}. We did not ask about personal privacy or collect personal information of annotators in the annotation process. We hired five annotators and payed each annotator \$0.05 and \$0.1 for annotating each story in ROC and WP, respectively. We decided the payment according to the average story length of two datasets. 
We admit that there may be still unpredictable bias in \textsc{mans} even though we have asked three experts to review all the annotated stories. 

Besides, we selected or constructed the test examples in \textsc{autos} based on general linguistic features. We did not adopt any selecting strategies or perturbation techniques which may introduce extra bias into \textsc{autos}.

\section*{Acknowledgments}
This work was supported by National Key R\&D Program of China, under Grant No. 2020AAA0104500. This work was jointly supported by the NSFC projects (Key project with No. 61936010 and regular project with No. 61876096), and the Guoqiang Institute of Tsinghua University, with Grant No. 2019GQG1. We thank THUNUS NExT Joint-Lab for the support. We would also like to thank the anonymous reviewers for their invaluable suggestions and feedback.

\bibliographystyle{acl_natbib}
\bibliography{anthology,acl2021}

\appendix
\section{Details for \textsc{mans}}
\subsection{Story Collection}
\paragraph{Data Processing} We collect machine-generated stories based on ROC and WP. To achieve better generation and generalization performance, we follow \citet{guan2020knowledge} to delexicalize stories in ROC by masking all the names with placeholders, and retain about 250 words (with correct sentence boundary) from the beginning and truncated the rest in WP.
\paragraph{Story Generation} After training, we use the generation models to generate stories based on the test sets of ROC and WP. We adopt nucleus sampling~\cite{holtzman2019curious} with $p=0.9$ for story generation to avoid as many repetition errors as possible, 
since such cases are easier for recognition and simulation~(we cover the repetition errors mainly with the test examples in \textsc{autos}).

\subsection{Manual Annotation}
Table~\ref{tab:error_case} shows examples of four error types and corresponding deduction points. The error types includes repetitive plots, unrelated events, conflicting logic and chaotic scenes.

\begin{table}[!ht]
\scriptsize
    \centering
    \begin{tabular}{p{134pt}p{61pt}}
    \toprule
    \textbf{Input: } My dad likes watermelon very much.&\textbf{Error Types}\\
    \midrule
    My mom bought him a watermelon. He \textit{picked it up and picked it up}. He \textit{picked it up and picked it up}. Now he have a big smile on his face.&\textbf{Repetitive plots}~(-1)\\
    \midrule
    He \textit{ate sweets and sweets}. \textit{He ate it all day}. \textit{He ate it all}. \textit{He ate it all}.&\textbf{Repetitive plots}~(-2)\\
    \midrule
    One day he decided to try a new \textit{recipe}. He bought all the \textit{ingredients}. He followed the \textit{recipe}. It was the best \textit{sauce} he ever tasted.&\textbf{Unrelated events} to the beginning~(-1)\\
    \midrule
    He decided to buy a \textit{banana}. He picked up a big \textit{oak tree}. He put it in the kitchen. He is happy with the watermelon.&\textbf{Unrelated events} to the beginning and within its context~(-2)\\
    \midrule
    He had a watermelon this morning. \textit{He wanted another one}. He went to buy one. \textit{He didn't want to eat watermelons}.&\textbf{Conflicting logic}~(-1)\\
    \midrule
    I buy a watermelon for him. It is pretty great for my dad. \textit{He doesn't like it}. He finally asked me to \textit{be his girlfriend}.&\textbf{Conflicting logic}~(-2)\\
    \midrule
    \textit{I had a watermelon when I was a child. I was feeding him fruits. I picked it up and put it in the house. He asked me to be his son.}&\textbf{Chaotic scenes}~(-2) \\
    \bottomrule
    \end{tabular}
    \caption{Examples of four error types and corresponding deduction points~(in the parentheses) given the same input. \textit{Italic} words indicate the keywords crucial for the errors.}
    \label{tab:error_case}
\end{table}

\subsection{Statistics}
\begin{figure}[t]
\centering
\subfigure{\includegraphics[width=\linewidth]{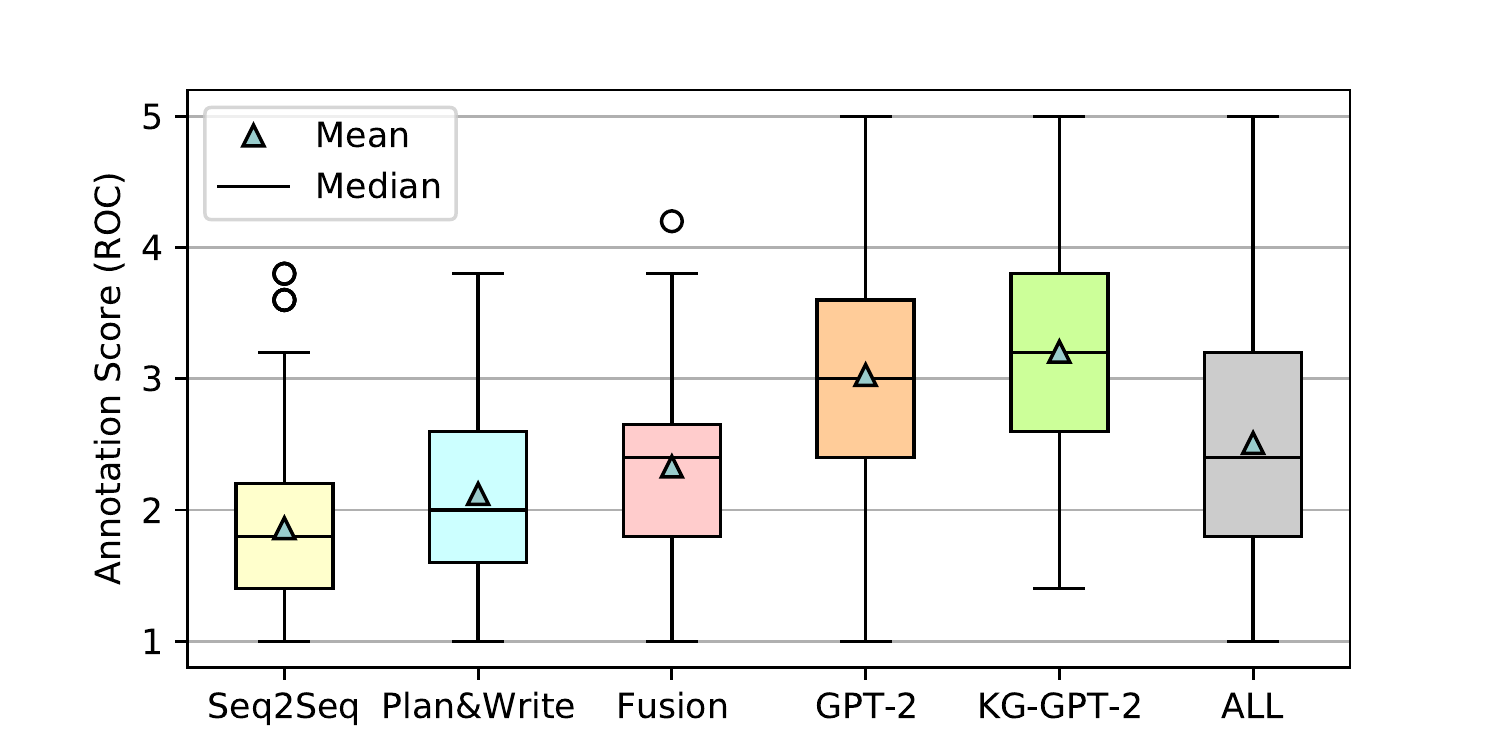}}\\
\vspace{-8mm}
\subfigure{\includegraphics[width=\linewidth]{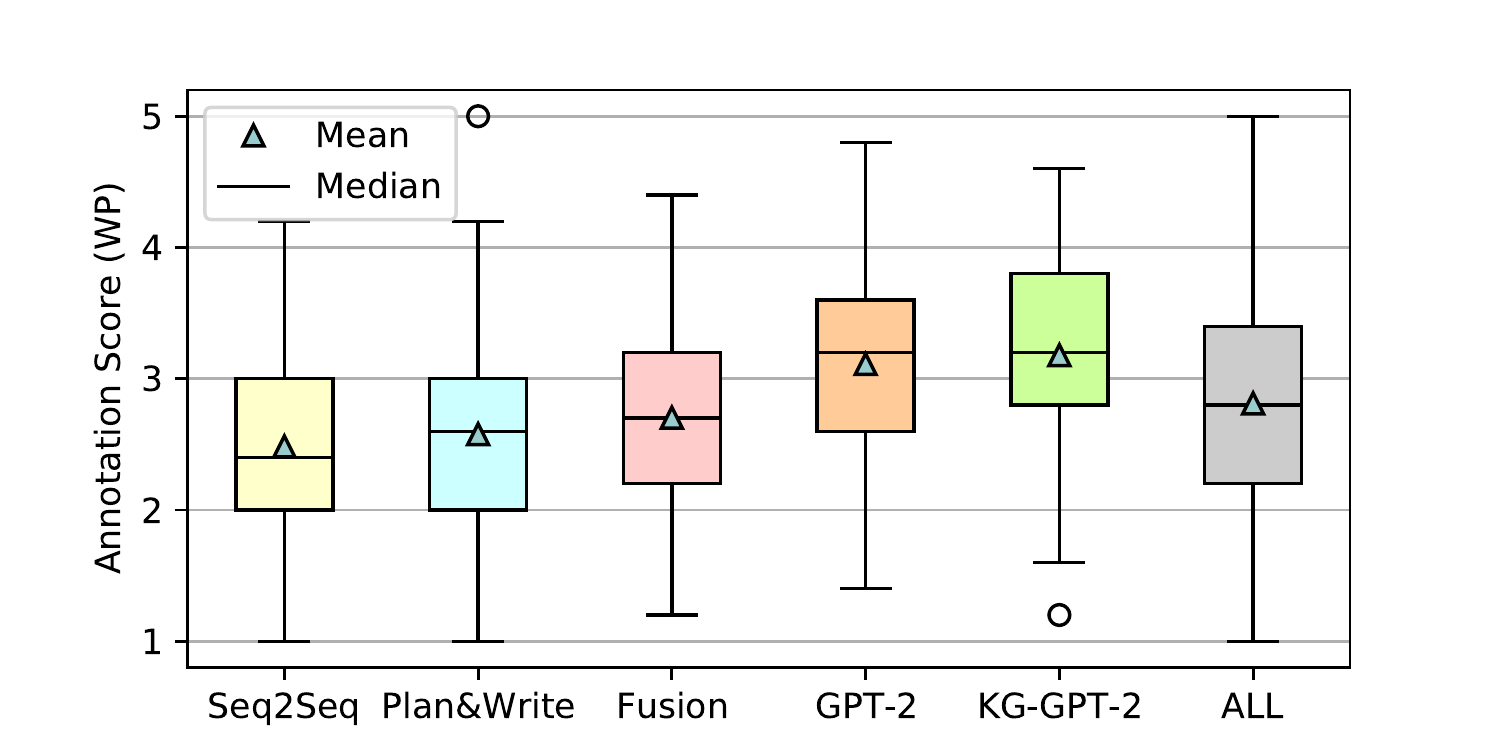}}
\vspace{-7mm}
\caption{Boxplot of human judgments for each story source (Top: ROC, Bottom: WP).}
\label{fig:quality}
\end{figure}

The Krippendorff’s $\alpha$ is 0.77 for ROC and 0.71 for WP, indicating a moderate inter-annotator agreement according to the interpretation in Table~\ref{tab:alpha}. 
We present the distribution of human judgments for different models in Figure~\ref{fig:quality} and other statistics in Table~\ref{tab:mags}. 
The results show the diversity of the stories in length and quality. 

\begin{table}[!ht]
\small
    \centering
    \begin{tabular}{cp{150pt}}
    \toprule
    \textbf{$\alpha$}&\textbf{Interpretation}\\
    \midrule
    $<0.67$&not good\\
    \midrule
    $0.67\sim0.8$&allowing tentative conclusions to be drawn\\
    \midrule
    $>0.8$&good reliability\\
    \bottomrule
    \end{tabular}
    \caption{Interpretation of Krippendorff's $\alpha$.}
    \label{tab:alpha}
\end{table}

\begin{table}[!ht]
\small
    \centering
    \begin{tabular}{lcc}
    \toprule
    \textbf{Statistics}&\textbf{ROC}&\textbf{WP}\\
    \midrule
    \textbf{Unique Inputs} & 200 & 200\\
    \textbf{Generated Stories (per Input)} & 5 & 5  \\
    \textbf{Generated Stories (totally)} & 1,000 & 1,000 \\    
    \textbf{Average Input Tokens}&9.26&22.09\\
    \textbf{Average Reference Tokens}&39.54&238.51\\ 
    \textbf{Average Story Tokens}&39.38&232.51\\
    
    \bottomrule
    \end{tabular}
    \caption{Statistics of \textsc{mans}. Text is tokenized with spaCy tokenizer\protect\footnotemark.}
    \label{tab:mags}
\end{table}
\footnotetext{\url{https://spacy.io/api/tokenizer}}

\subsection{Correlation with Human Judgments}
We experimented with more popular metrics as follows: \texttt{ROUGE-L}~\cite{lin-2004-rouge}, \texttt{METEOR}~\cite{banerjee2005meteor}, embedding based metrics~(including \texttt{Greedy Matching}, \texttt{Embedding Average} and \texttt{Vector extrema}~\cite{liu2016not}) with BERT embedding, \texttt{BERTScore} (inlcuding \texttt{P}recision and \texttt{R}ecall), and \texttt{MoverScore}. \texttt{RUBER}, and the supervised metric \texttt{BLEURT} which is fine-tuned on the released annotation results from \citet{guan2020union}. The experiment results is shown in Table~\ref{tab:corr_more}. 

\begin{table}[!ht]
\small
    \centering
    \begin{tabular}{lll}
    \toprule
{\textbf{Metrics}} &\textbf{ROC}&\textbf{WP}\\    
	\midrule
	\midrule
	\multicolumn{1}{l}{\textbf{Referenced Metrics}}\\
	\midrule
    \texttt{BLEU}&-0.0239&-0.0537\\
    \texttt{ROUGE-L}&~0.0188&-0.0107\\
    \texttt{METEOR}&~0.0155&-0.0079\\
	\texttt{Greedy Matching}&~\textbf{0.1892}$^{*}$&-0.0510\\
	\texttt{Vector Average}&~0.1840$^{*}$&-0.0429\\
	\texttt{Vector Extrema}&~0.1021$^{*}$&-0.0241\\
	\texttt{BERTScore-P}&~0.1538$^{*}$&~0.0857$^{*}$\\
	\texttt{BERTScore-R}&~0.0838$^{*}$&-0.0215\\
    \texttt{BERTScore-F1}&~0.1271$^{*}$&~0.0329\\
	\texttt{MoverScore}&~0.1294$^{*}$&-0.0586\\
	\texttt{R$_r$-BERT}&~0.0808$^{*}$&~\textbf{0.1567}$^{*}$\\
		\midrule
	\midrule
	\multicolumn{1}{l}{\textbf{Unreferenced Metrics}}\\
	\midrule		
	\texttt{PPL (P)}&~0.2547$^{*}$&~0.3033$^{*}$\\
	\texttt{PPL (F)}&~0.2817$^{*}$&~0.2952$^{*}$\\
	\texttt{R$_u$-BERT}&~0.0830$^{*}$&~0.1666$^{*}$\\
    {\texttt{\textsc{Union}}}&~\textbf{0.4119}$^{*}$&~\textbf{0.3256}$^{*}$\\
	\midrule
	\midrule
	\multicolumn{1}{l}{\textbf{Hybrid Metrics}}\\
	\midrule	
	\texttt{RUBER}&~0.0119&-0.0527\\
    \texttt{RUBER-BERT}&~{0.1434}$^{*}$&~\textbf{0.2116}$^{*}$\\
    \texttt{BLEURT}&~\textbf{0.3163}$^{*}$&~{0.1738}$^{*}$\\
	\bottomrule
    \end{tabular}
    \caption{Pearson correlation with human judgments on \textsc{mans}. The best performance for each type of metrics is highlighted in \textbf{bold}. The correlation scores marked with * indicate the result significantly correlates with human judgments (p-value<0.01).}
    \label{tab:corr_more}
\end{table}

\section{Details for \textsc{autos}}

\begin{table*}[!ht]
\scriptsize
    \centering
    \begin{tabular}{p{55pt}p{170pt}p{194pt}}
    \toprule
    \textbf{Types}&\textbf{Conjunction, Preposition, Adverb}&\textbf{Noun, Verb, Adjective}\\
    \midrule
    \textbf{Negated}&no, not, never, neither, hardly, unlikely, rarely, seldom, impatiently, uncertainly (incomplete listing, 215 in total)&none, nobody, nothing, disable, disagree, disappear, illegal, inability, inactive, unhappy, unfortunately (incomplete listing, 164 in total)\\
    \midrule
    \textbf{Causality-related}&so, because, since, therefore, why&cause, reason, result, effect, purpose, aim, sake, consequence, causal\\
    \midrule
    \textbf{Time-related}&after, before, previously, simultaneously, currently, meanwhile, then, now, ever, again, once, anytime, when, while, never, always, usually, often, sometimes, usually, early, lately, already, forever, ago, yesterday, today, tomorrow&ending, beginning, previous, simultaneous, current, temporary, contemporary, temporal, second, minute, hour, day, month, year, century, past, future, present, delay, night, evening, morning, afternoon, noon, morning\\
    \bottomrule
    \end{tabular}
    \caption{Negated words, causality-related words, time-related words which are used to create test examples within the aspects ``Consistency'', ``Causal Relationship'' and ``Temporal Relationship'', respectively.}
    \label{tab:caus_time}
\end{table*}

\subsection{Construction}
We list some technical details for constructing \textsc{autos} within different aspects as follows:
\begin{itemize}
\item \textbf{Semantic Repetition} and \textbf{Paraphrases}: We present several examples for paraphrase generation in Table~\ref{tab:bt}. We adopt MoverScore and BLEU-1 to measure the semantic similarity and word overlap between the paraphrases and the original sentences, respectively. We finally only use the paraphrase whose MoverScore is larger than 0.4 and BLEU-1 is less than 0.6 with the original sentence, because they achieve both high semantic similarity and low word overlap.
\item \textbf{Character Behaviour}: We recognize the personal pronouns in a story following Table~\ref{tab:pron}. We select those stories which contain at least three types of person~(i.e., at least three pronouns from different rows) as the coherent examples. And when substituting the pronouns to create incoherent examples, we only perform the substitution in the same column~(e.g., \textit{``my''} can be only substituted 
with \textit{``our''}, \textit{``your''}, etc.) for better grammaticality.
\item \textbf{Consistency}, \textbf{Causal} \textbf{and Temporal Relationship}: We present the negated words, causality-related words and the time-related words in Table~\ref{tab:caus_time}.
\end{itemize}

\begin{table}[!ht]
\small
    \centering
    \begin{tabular}{lllllll}
    \toprule
\textbf{Subj}&\textbf{Obj}&\textbf{Poss~(A)}&\textbf{Poss~(N)}&\textbf{Ref}\\
\midrule
i&me&my&mine&myself\\
we&us&our&ours&ourselves\\
you&you&your&yours&yourself\\
you&you&your&yours&yourselves\\
he&him&his&his&himself\\
she&her&her&hers&herself\\
it&it&its&its&itself\\
they&them&their&theirs&themselves\\
	\bottomrule
    \end{tabular}
    \caption{Personal pronouns which are used to create test examples within the aspect ``Character Behaviour''. Each row specifies one type of person, which has five forms: \textbf{subj}ective pronouns, \textbf{obj}ective pronouns, \textbf{poss}essive \textbf{a}djectives, \textbf{poss}essive \textbf{n}ouns and \textbf{ref}lexive pronouns.}
    \label{tab:pron}
\end{table}

\begin{table}[!ht]
\scriptsize
    \centering
    \begin{tabular}{p{75pt}p{70pt}p{12pt}p{12pt}}
    \toprule
\textbf{Original Sentences}&\textbf{Paraphrases}&\textbf{M}&\textbf{B}\\
\midrule
I filled it with the sodas.& I put music into the world and enjoy it.&0.05&0.40\\
\midrule
He went several more miles out of his way. & He has made kilometers more.&0.16&0.26\\
\midrule
She screamed loudly to attract the attention of her audience.& She yelled out loud for the attention of the public.&0.42&0.45\\
\midrule
He hired an attorney.& He employed a lawyer.&0.57&0.40\\
\midrule
She watched a video of the play later. & She later watched a video of the play. &0.75 &0.89\\
	\bottomrule
    \end{tabular}
    \caption{Examples for paraphrase generation. \textbf{M} and \textbf{B} mean the MoverScore and BLEU-1 between the paraphrases and the original sentences, respectively.}
    \label{tab:bt}
\end{table}

\begin{table}[!ht]
\scriptsize
    \centering
    \begin{tabular}{p{183pt}p{12pt}}
    \toprule
\textbf{Cases}&\textbf{S}\\
\midrule
1. She \textit{head} to the city. & 0.07\\
\midrule
2. A strange elderly woman \textit{and} called his name. & 0.20\\
\midrule
3. They walked home several more times \textit{whenever that}.&0.41\\
\midrule
\midrule
4. One day Mary needed to leave the airport . She had no idea on how to get a taxi though. Asking for some help she learned about lyft. \underline{She had no idea how to get a taxi}. Within a hour she was at home, happy with her decision.& 0.66\\
\midrule
5. Jack was invited to a holiday party. He wanted to bring his hostess a gift. But he had no clue what! \underline{Before} googling, he decided on a bottle of wine . his hostess was very pleased with it.&0.95\\
	\bottomrule
    \end{tabular}
    \caption{Examples for the grammaticality classifier. The examples are sentences or stories selected from the incoherent examples of the discrimination test set of \textsc{autos}. \textbf{S} means the classifier score$\in[0,1]$~(1 is the best). The \textit{italic} words are ungrammatical, and the \underline{underlined} ones are unreasonable in logic but grammatical.}
    \label{tab:cls}
\end{table}

\begin{table}[!ht]
\small
    \centering
    \begin{tabular}{l|cc|cc}
    \toprule
    \multirow{2}{*}{\textbf{Datasets}}&\multicolumn{2}{c|}{\textbf{Coherent}}&\multicolumn{2}{c}{\textbf{Incoherent}}\\
    &\textbf{Input}&\textbf{Story}&\textbf{Input}&\textbf{Story}\\
    \midrule
    \textbf{ROC}&8.76&39.28&8.88&40.39\\
    \textbf{WP}&30.02&235.83&30.28&228.04\\
    \bottomrule
    \end{tabular}
    \caption{Statistics of the discrimination test set in \textsc{autos}.  \textbf{Input} and \textbf{Story} is the average number of tokens in the inputs and stories. \textbf{Coherent} means the coherent examples which are selected from the human-written stories. \textbf{Incoherent} means the incoherent examples which are automatically constructed by perturbing the human-written stories. }
    \label{tab:detail_dis_stat}
\end{table}

\begin{table}[!ht]
\small
    \centering
    \begin{tabular}{l|cc|cc}
    \toprule
    \multirow{2}{*}{\textbf{Datasets}}&\multicolumn{2}{c|}{\textbf{Human}}&\multicolumn{2}{c}{\textbf{Dis}}\\
    &\textbf{Input}&\textbf{Story}&\textbf{Input}&\textbf{Story}\\ 
    \midrule
    \textbf{ROC}&9.03&40.66&9.23&40.57\\
    \textbf{WP}&29.65&211.19&29.60&234.40\\
    \bottomrule
    \end{tabular}
    \caption{Statistics of the invariance test set in \textsc{autos}.  \textbf{Input} and \textbf{Story} is the average number of tokens in the inputs and stories. \textbf{Human} and \textbf{Dis} means the human-written coherent stories and incoherent samples (sampled from the discrimination test set) to be perturbed, respectively.
    }
    \label{tab:detail_inv_stat}
\end{table}

\subsection{Grammaticality Classifier}
We train a binary classifier on the CoLA corpus~\cite{warstadt2019neural} to learn to judge the grammaticality, and then filter out those examples that are classified as ungrammatical~(the classifier score less than 0.5). For simplicity, we directly use the public model from TextAttack~\cite{morris2020textattack} as a classifier to filter out those examples in \textsc{autos} with poor grammaticality. The classifier is fine-tuned on the CoLA corpus based on BERT and achieves an accuracy of 82.90\% on the test set of CoLA. Furthermore, if we suppose that all of the human-written stories in ROC and WP are grammatical, the accuracy of the classifier on the stories would be 96.48\% and 65.68\% for ROC and WP, respectively. The results are intuitive since stories in WP may contain much informal English~(e.g., website link). We present several examples in Table~\ref{tab:cls} to further indicate the usefulness of the classifier. We can see that the classifier can detect the grammar errors in multiple aspects such as verb forms~(e.g., \textit{``head''} should be \textit{``heads''} for case 1) and sentence elements~(e.g., the predicate is missing for case 3). And the classifier would give the grammatical sentences high scores although they may be unreasonable in logic~(e.g., repetitive texts for case 4 and conflicting plot for case 5). Finally, we filter out about 21.69\% and 50.15\% examples for ROC/WP, respectively.

\subsection{Statistics}
We show the statistics of the discrimination test set and the invariance test set in \textsc{autos} in Table~\ref{tab:detail_dis_stat} and Table~\ref{tab:detail_inv_stat}, respectively.

\end{document}